%% file: Trail2023_short.tex
\documentclass[runningheads,a4paper]{llncs}

\usepackage{ifxetex}
\ifxetex%
	\usepackage{fontspec}
	\usepackage{polyglossia}
	\setmainlanguage{english}
\else
	\usepackage[utf8]{inputenc}
	\usepackage[T1]{fontenc}
	\usepackage[english]{babel}
\fi

\usepackage{amsmath}
\usepackage{amssymb}
\usepackage{url}
\usepackage{float}
\usepackage{titling}
\usepackage{wrapfig}
\usepackage{booktabs}
\usepackage{csquotes}
\usepackage{fancyhdr}
\usepackage{graphicx}
\usepackage{lastpage}
\usepackage[all]{nowidow}
\usepackage[inline]{enumitem}
\usepackage[usenames,dvipsnames]{xcolor}

\usepackage{varioref}
\usepackage[hidelinks]{hyperref}
\usepackage[noabbrev,nameinlink]{cleveref}

\usepackage{lipsum}

\input{macros}
\title{TRAIL Team Description Paper for RoboCup@Home 2023}

\author{
    Chikaha Tsuji,
    Dai Komukai,
    Mimo Shirasaka,
    Hikaru Wada, \\
    Tsunekazu Omija,
    Aoi Horo,
    Daiki Furuta,
    Saki Yamaguchi,\\
    So Ikoma,
    Soshi Tsunashima,
    Masato Kobayashi,
    Koki Ishimoto,\\
    Yuya Ikeda,
    Tatsuya Matsushima,
    Yusuke Iwasawa,
    Yutaka Matsuo\\
    \textit{The University of Tokyo}\\
    \url{https://trail.t.u-tokyo.ac.jp/project/robocup2023/}}
\institute{Affiliation name and address, \\
\texttt{http://devoted-web-site.url}}

\date{December 4, 2022}

\begin{document}
\maketitle

\begin{abstract}

Our team, TRAIL, consists of AI/ML laboratory members from The University of Tokyo. We leverage our extensive research experience in state-of-the-art machine learning to build general-purpose in-home service robots. We previously participated in two competitions using Human Support Robot (HSR): RoboCup@Home Japan Open 2020 (DSPL) and World Robot Summit 2020, equivalent to RoboCup World Tournament. Throughout the competitions, we showed that a data-driven approach is effective for performing in-home tasks. Aiming for further development of building a versatile and fast-adaptable system, in RoboCup @Home 2023, we unify three technologies that have recently been evaluated as components in the fields of deep learning and robot learning into a real household robot system. In addition, to stimulate research all over the RoboCup@Home community, we build a platform that manages data collected from each site belonging to the community around the world, taking advantage of the characteristics of the community.
\end{abstract}

\vspace{-0.5\baselineskip}
\section{Introduction and Relevance}
\subsection{TRAIL}
Our team, TRAIL (Tokyo Robotics and AI Lab), was launched in 2020 as a project and was organized in 2021 as a group under Matsuo Laboratory\footnote{\url{https://weblab.t.u-tokyo.ac.jp/en/}} at The University of Tokyo.
Our laboratory mainly engages in fundamental research on deep learning, especially world models, and also focuses on robotics as an application of deep learning.
TRAIL aims to realize general-purpose robots that can perform a wide variety of tasks in diverse environments by utilizing robot learning.
As part of these activities, we are making the most of our knowledge as an AI/ML lab to conduct experiments to build a real robot system that can carry out daily life support tasks in household environments. More specifically, we aim to build a well-generalized and fast-adaptable system leveraging technologies such as \textit{foundation model, Sim2Real} and \textit{imitation learning}.

\subsection{The Experience and Achievements in Local Tournaments}

So far, we have participated in the competitions listed in Table~\ref{table:recent_competition} as opportunities to validate our in-home service robot system and won awards as indicated. In the competitions, we leveraged a data-driven approach to address tidy-up tasks in household environments rather than a pre-programmed approach to handle numerous edge cases. Through the competitions, we showed that a data-driven approach adapts better to diverse household environments and flexibly handles various edge cases than a pre-programmed approach; refer to~\cite{doi:10.1080/01691864.2022.2114297} for details. However, while a tidy-up task is an essential aspect of in-home service robots, there are other components, such as human-robot interaction, safe and flexible manipulation and navigation in a more dynamic environment, which were not included in the competitions we participated in. Fortunately, these elements are included in RoboCup@Home 2023, which is suitable for realizing our ends. Therefore, we plan to extend our current robot system toward the realization of a general-purpose in-home service robot, and we intend to demonstrate the achievements of our activities in RoboCup@Home 2023.
\begin{table}[tb]
\caption{\small Results of Competitions}
  \vspace{-0.6\baselineskip}
  \label{table:recent_competition}
  \centering
    \scalebox{0.85}{
  \begin{tabular}{c|c}
    \hline
       Competitions &  Results \\
    \hline \hline
    RoboCup@Home Japan Open 2020  & 2nd place at DSPL league\\
     & 1st place at Technical Challenge \\
    \hline
    WRS2020, equivalent to RoboCup World Tournament  &  2nd place at Partner Robot Challenge,\\
     & equivanlent to DSPL league\\
    \hline
  \end{tabular}
}
\end{table}

\begin{figure}[t]
	\centering
	\includegraphics[width=0.9\textwidth]{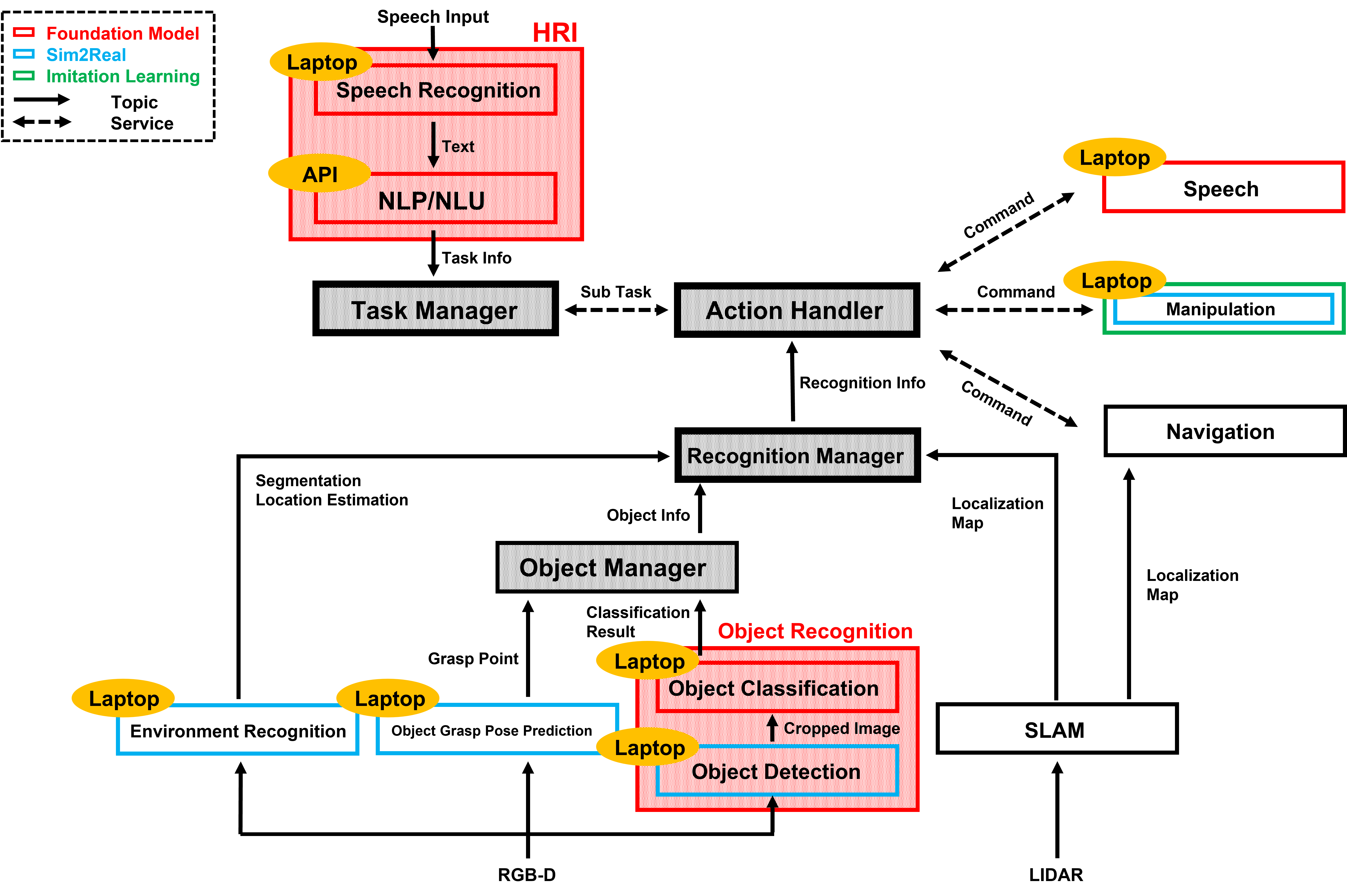}
\caption{\small The configuration of our system using the three technologies: \textit{foundation model} (red-framed) adapted instantly with prompt tuning (red-highlighted), \textit{Sim2Real} (blue-framed), and \textit{imitation learning} (green-framed).}
	\label{fig:System_Overview}
\end{figure}

\vspace{-0.5\baselineskip}
\section{Approach}

We aim to build a \textit{foundation model}-centric system that is versatile and fast-adaptable. To build this system, we focus on three key technologies: \textit{foundation model}, \textit{Sim2Real}, and \textit{imitation learning}. Fig.~\ref{fig:System_Overview} visualizes how the three key technologies are utilized in each module. In the following sections, 2.3, 2.4, and 2.5,  task-level applications of the three key technologies are described, respectively.

\subsection{Background}
In RoboCup@Home competitions, same as in actual diverse household environments, participants must cope with both predefined settings announced on the setup days, and uncertainties revealed not until the competition days. We could use deep neural network (DNN)-based models specialized for specific tasks, however, they require high costs of collecting data and annotating in terms of time and effort when fine-tuning the models to adapt to new environments.

Recently,  \textit{foundation model} has been shown to be highly generalizable in the field of deep learning~\cite{bommasani2021opportunities}, especially in image and natural language processing.
Besides, in the robot learning area, \textit{Sim2Real} and \textit{imitation learning} have been shown as components to provide flexible, versatile, and fast-adaptable recognition and control. We aim to solve the challenges in RoboCup@Home by leveraging the three key technologies promising to build an entire system that is well-generalized and fast-adaptable.

We have found out that there are mainly two challenges in unifying the three technologies into an actual robot system; one is to operate a system with large-scale models, such as \textit{foundation models}, in real-time, while another is the obscurity to what extent the remarkable technologies as above can be applied in practice.
Thus, through RoboCup@Home 2023, we aim to build a system with large-scale models that can operate in real-time and seek the best way to integrate the technologies with our robot system.

\subsection{System Overview}

Fig.~\ref{fig:System_Overview} shows the configuration overview of our \textit{foundation model}-centric system using the three key technologies; \textit{foundation model}, \textit{Sim2Real}, and \textit{imitation learning}.
First, task instructions by natural language speech inputs are processed in Human-Robot Interaction (HRI) module using \textit{foundation models} (Section 2.3.2), and Task Manager subdivides the tasks based on the processed linguistic information.
Next, integrating the environment-specific information obtained from sensor data, Action Handler sends commands to the actuator modules, such as Speech and Manipulation (Section 2.5) modules, which also leverage the three key technologies. Sensor data is processed by modules utilizing \textit{foundation model} and \textit{Sim2Real}; Object Recognition (Sections 2.3.1 and 2.4.3), Object Grasp Pose Prediction (Section 2.4.2), and Environment Recognition (Section 2.4.1). Thereby, the system can be sufficiently generalized to handle the uncertainties during the competition. In addition, the system can adapt quickly to new environments on setup and competition days, as represented by prompt tuning, which is another advantage of \textit{foundation models}.

\subsection{Foundation Model}
\textit{Foundation model} is a large-scale model that is pre-trained with a large amount and a variety of data. By engineering the prompts, which are given in natural language to specify the task, it can handle various downstream tasks and edge cases even without additional learning.

\subsubsection{Foundation Model for Object Recognition}\mbox{}\\

\begin{figure}[tb]
  \begin{minipage}{0.59\hsize}
    \begin{center}
      \scalebox{0.19}{
        \includegraphics{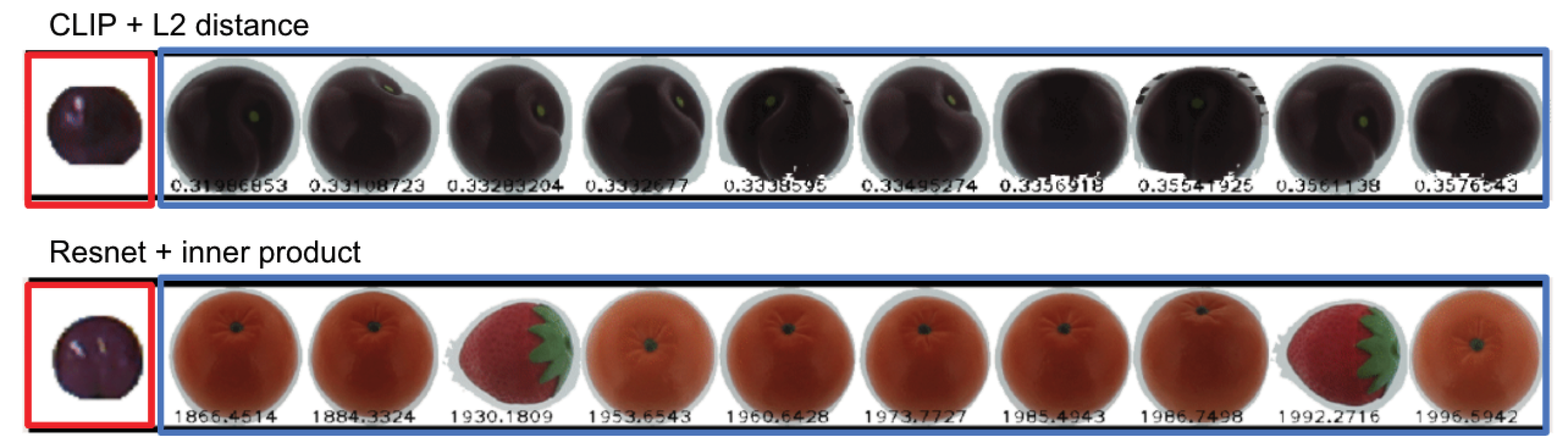}}
      {\begin{center} (a) CLIP (upper) and Resnet18 (lower)\end{center}}
    \end{center}
  \end{minipage}
  \begin{minipage}{0.4\hsize}
    \begin{center}
      \scalebox{0.15}{
        \includegraphics{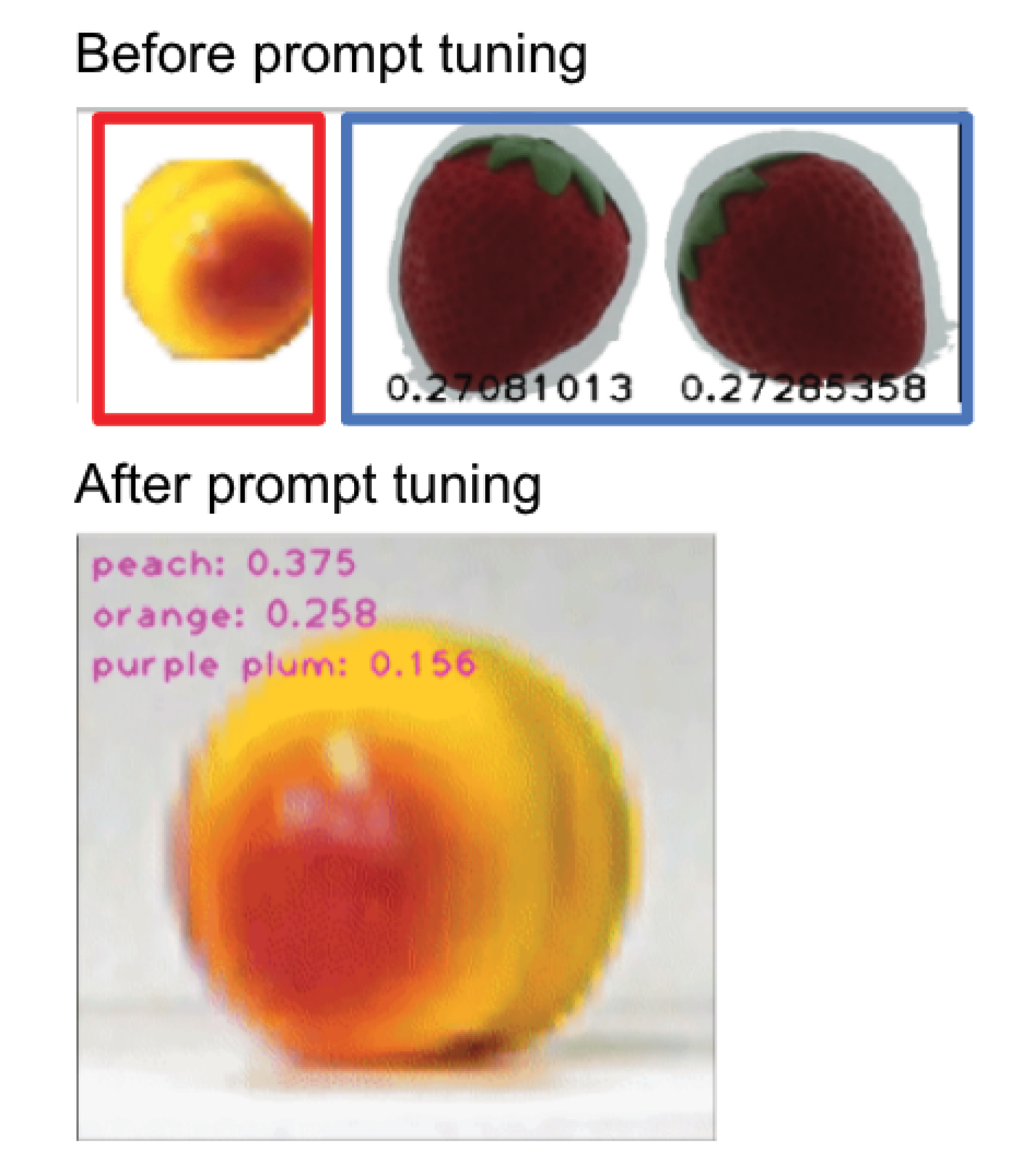}}
      {\begin{center} (b)Before and after prompt tuning\end{center}}
    \end{center}
  \end{minipage}
        \caption{\small Results of object classification using CLIP. (a) CLIP and Resnet18 results. The red-framed image is the recognized image, and the blue-framed images are images classified to be similar. (b) Results before and after prompt tuning. With prompt tuning, the model changes recognition results from strawberry~(wrong) to peach~(correct). }
        \label{fig:object_classification}
\end{figure}

Object Recognition module consists of two modules: Object Detection module and Object Classification module. In Object Detection module, we combine Mask R-CNN~\cite{he2017mask}
and UOIS~\cite{UOIS} to generate segmentation images that extract detected objects. In our object classification module, we leverage CLIP~\cite{radford2021learning}
to classify objects using segmentation images from Object Detection module.
Here, CLIP is a \textit{foundation model} for image classification that inputs the prompts, the text description of classifying objects, and the images and then outputs the probabilities of identification between the name and the image.
As shown in Fig.~\ref{fig:object_classification}(a), the developed CLIP-based model has higher classification accuracy than the pre-trained ResNet18 model~(often used as a baseline). In addition, we succeeded in improving the accuracy by editing the prompts~(prompt-tuning) as in Fig.~\ref{fig:object_classification}(b).
We can expect the model to adapt quickly to predefined objects announced during the setup days right before the competition days.

\subsubsection{Foundation Model for Human-Robot-Interaciton}\mbox{}\\

There are many tasks in the RoboCup@Home competition that require various human-robot interactions via natural languages, such as question-and-answer, conversations, and even visual information descriptions.
However, it is not practical to tune a large language model for each downstream task and then store and use different models for each task.
Therefore, we leverage \textit{foundation models} that excel in dealing with natural language; Whisper\footnote{\url{https://github.com/openai/whisper}} for speach recognition and GPT-3 API\footnote{\url{https://beta.openai.com/docs/models/gpt-3}} for NLP/NLU.
Our speech recognition module uses an adjustable \textit{foundation model} so that we can utilize extracted feature values inside the model, unlike other APIs (e.g. Google Cloud Speech-to-Text API), which only provides output results.
In NLP/NLU module, we are now developing a system that shares a single language model for all tasks by using a prompt-tuning approach instead of switching between tuned models for each task.

\subsection{Sim2Real}
The light conditions and furniture positions may alter during the competition, as well as real household environments. Accordingly, we heavily use deep learning and simulators, especially in the essential technologies that form the basis of the system in all tasks, to make our system robust to such uncertainties and changing environments.

\subsubsection{Sim2Real for Environment Recognition}\mbox{}\\

\begin{figure}[tbp]
  \begin{minipage}[h]{0.49\linewidth}
    \centering
    \includegraphics[width=0.75\linewidth]{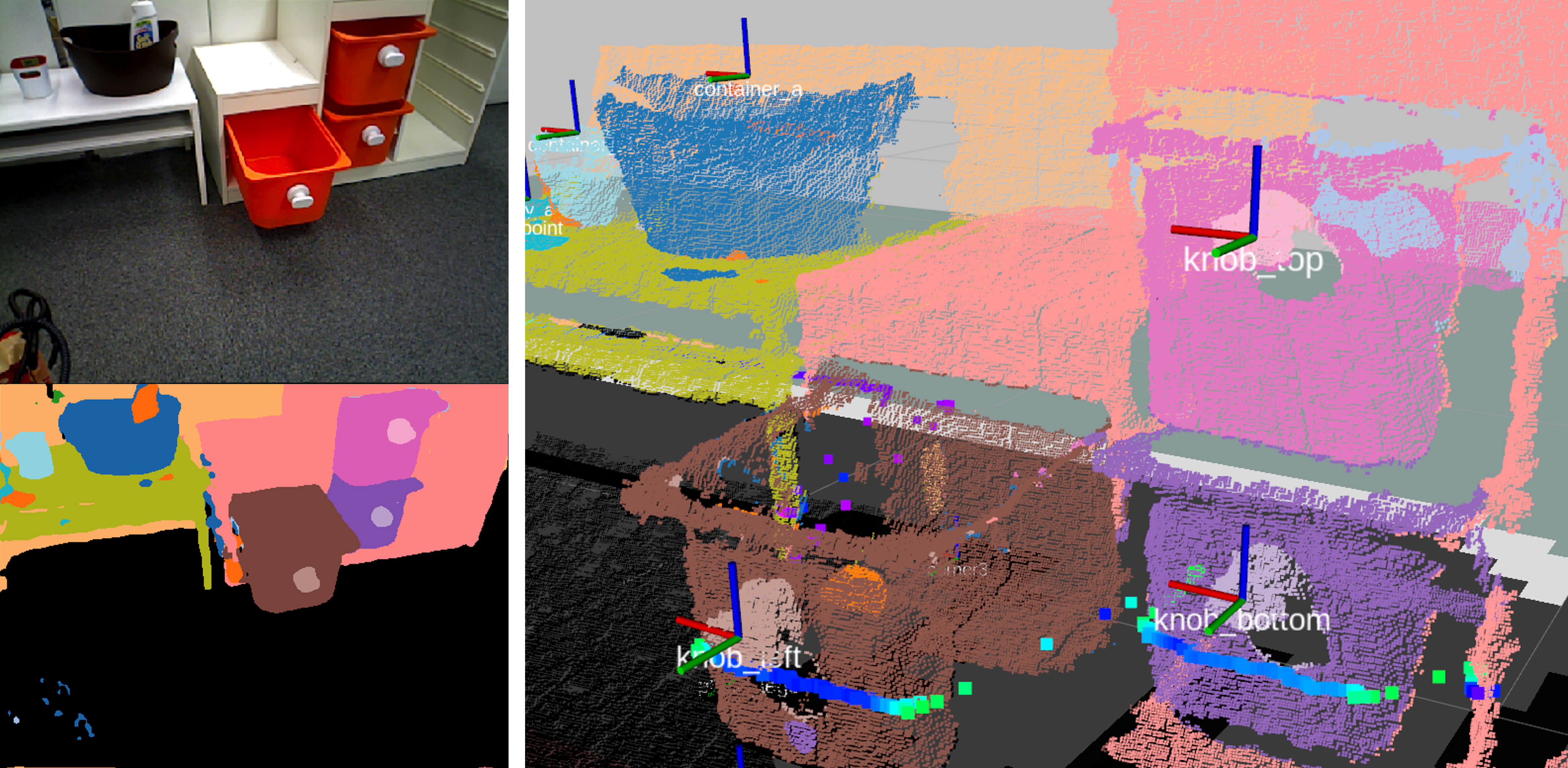}
    \caption{\small  Left: Segmentation results of drawer knobs in a real room. The model is robust enough to change in position and shape.
Right: Localization result of drawer knobs and containers using the segmentation results.}
    \label{fig:randomknobs}
  \end{minipage}
  \begin{minipage}[h]{0.49\linewidth}
    \centering
    \includegraphics[width=0.75\linewidth]{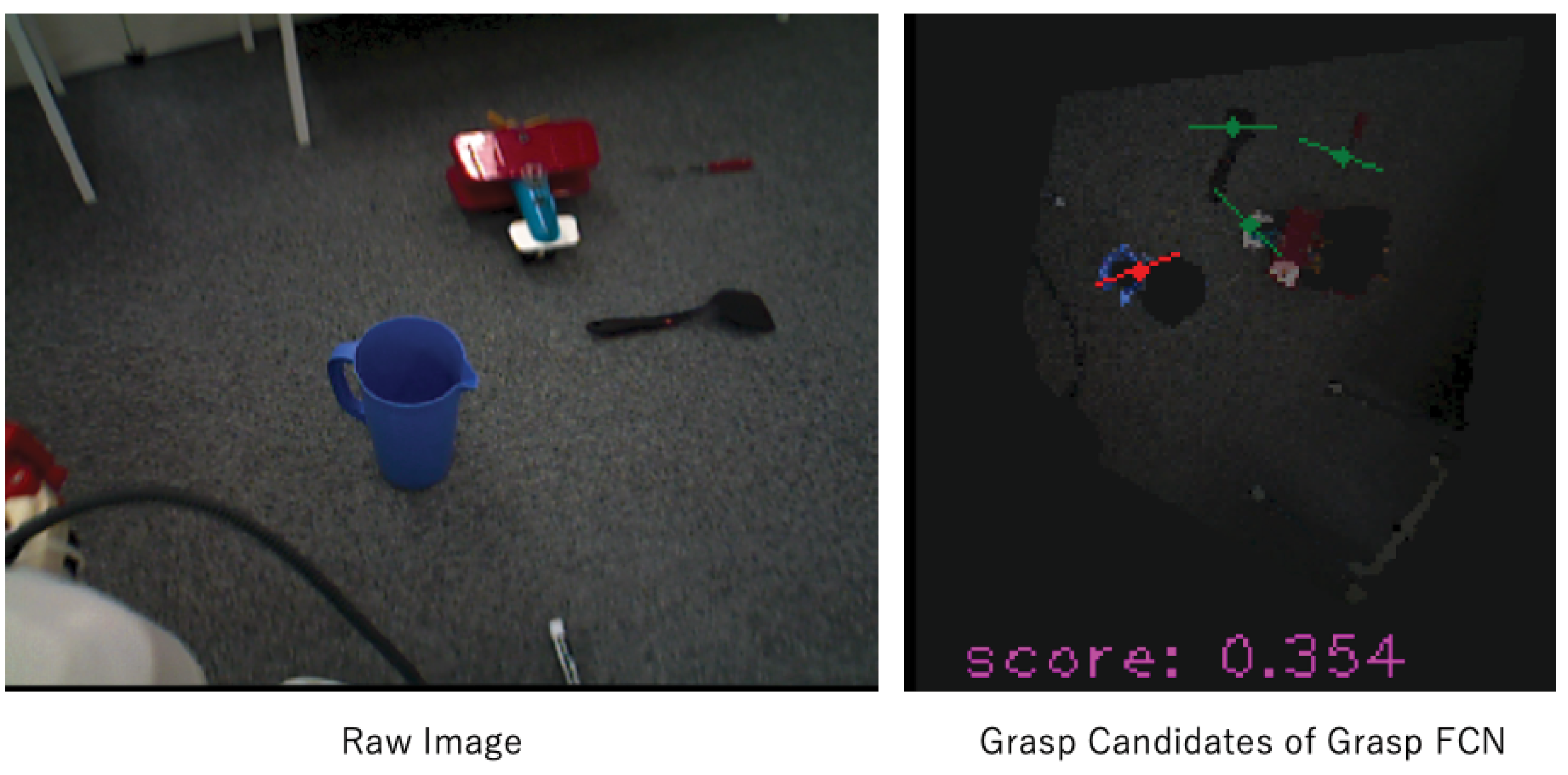}
    \caption{\small  Grasp FCN produces the grasp candidates with the graspability score.
Left: RGB image from the head camera.
Right: The red line marks the grasp candidate with the highest graspability score, and the greens mark other candidates. }
    \label{fig:FCN}
  \end{minipage}
\end{figure}

We trained a fully convolutional neural network (FCN)~\cite{long2015fully} to segment a given depth image.
The model inputs a depth image and then outputs the category for each pixel.
We collected data through a PyBullet\footnote{\url{https://github.com/bulletphysics/bullet3}} simulator ported from the existing task environment of a Gazebo simulator developed under the hsr-project\footnote{\url{https://github.com/hsr-project/tmc_wrs_gazebo}} in WRS. We leveraged domain randomization in terms of viewpoints, positions, rotations, and shapes of various objects and environmental changes.
In addition, we sampled meshes from ShapeNet~\cite{chang2015shapenet}
and spawned them all over the room. Considering the gap between the simulator and the real world, we intentionally did not use color information to ensure robustness to changes, especially in light conditions.
Fig.~\ref{fig:randomknobs} shows the segmentation results in a real room.

\subsubsection{Sim2Real for Object Grasp Pose Prediction}\mbox{}\\

While we use straightforward methods for grasping simple objects~(PCA on object point clouds), we take advantage of the learned policy trained on a simulator described in 2.3.1 for grasping more complex objects, such as bowls and ropes. We trained a grasp model based on FCN using the Deep Q-Learing method (Grasp FCN). This model inputs a heightmap generated by projecting point clouds onto the XY plane and outputs a map of Q values corresponding to the probability of a successful grasp at the pixel location.
Fig.~\ref{fig:FCN} shows the output result of Grasp FCN. The output Q-values are used for estimating not only the grasp pose but also the priority of the target object among many objects.

\subsubsection{Sim2Real for Object Recognition}\mbox{}\\

\begin{figure}[tbp]
  \begin{minipage}[h]{0.49\linewidth}
    \centering
    \includegraphics[width=\linewidth]{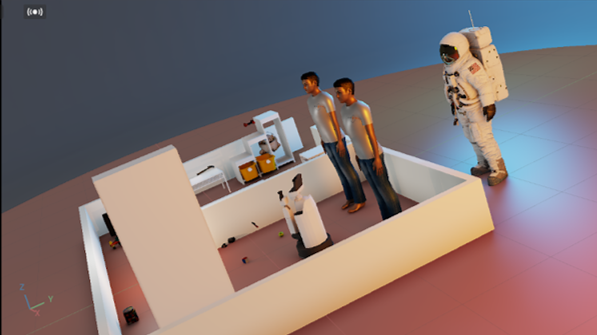}
    \caption{\small  The task environment in Isaac Sim, a photorealistic simulator.Linked to ROS, the simulator can be used for data generation and the robot's operation test.}
    \label{fig:isaacsim}
      \vspace{-12mm}
  \end{minipage}
  \begin{minipage}[h]{0.49\linewidth}
    \centering
    \includegraphics[width=\linewidth]{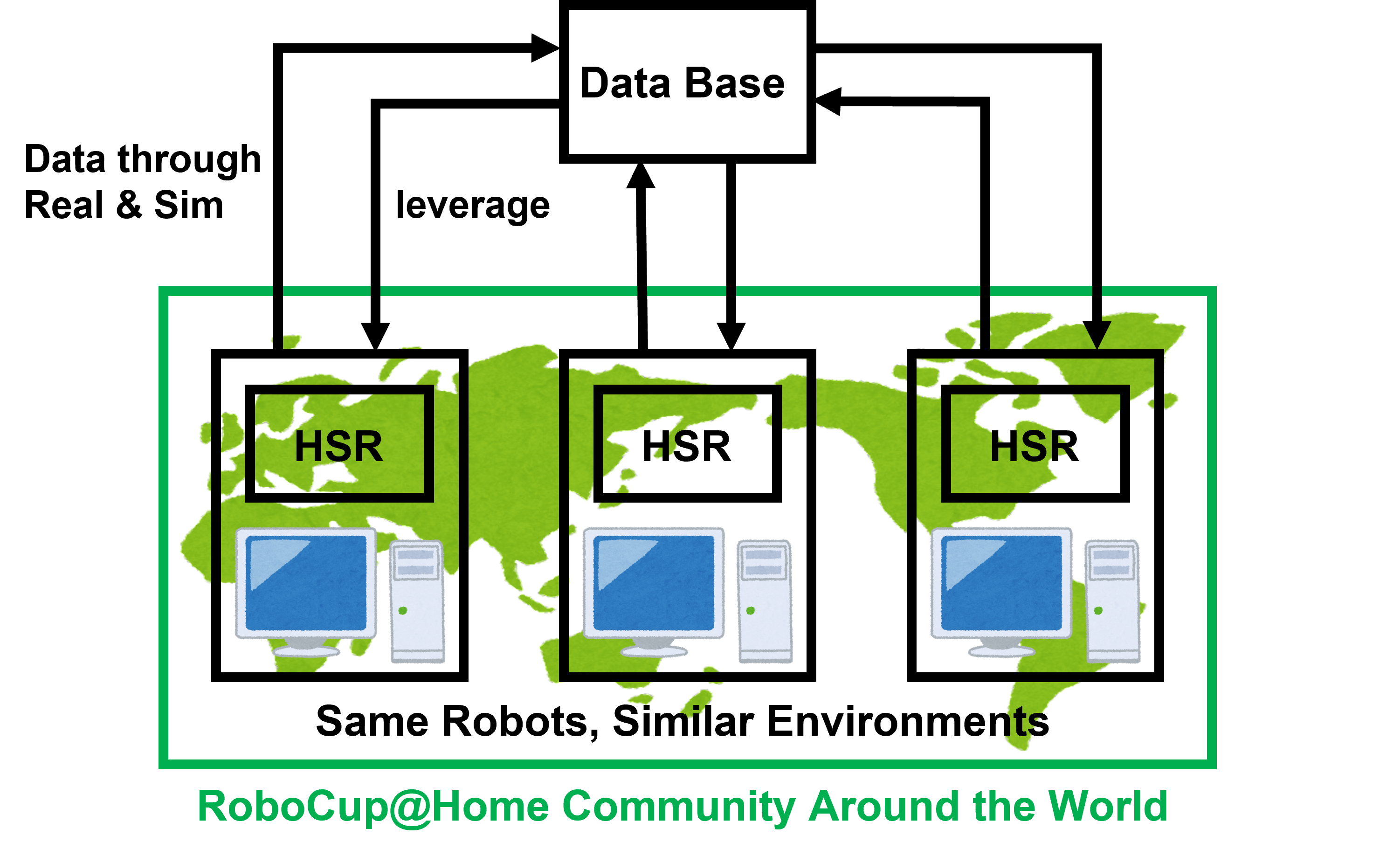}
    \caption{\small  Conceptual diagram of the platform. Taking advantage of the characteristics of the RoboCup@Home Community, we aim to build a system to collect and utilize data for robots.}
    \label{fig:Data}
  \end{minipage}
\end{figure}

Correct recognition of furniture and objects is crucial not only for object recognition and manipulation but also for safe navigation.
Training a generalized recognition model requires a large and diverse data set.
However, collecting data in the real world is costly in terms of time and effort, so simulators are expected to be used.
Unfortunately, the rendering of Gazebo and PyBullet simulators described in 2.3.1 has large domain gap from the real world, which makes it difficult to \textit{Sim2Real} transfer, especially in object recognition, where information on color, texture, and light conditions can be important factors.
Therefore, we utilize a more photorealistic and realistic simulator, Isaac Sim\footnote{\url{https://developer.nvidia.com/isaac-sim}}. We have already developed a task environment as in Fig.~\ref{fig:isaacsim}.
Using this environment, we are currently developing a generalized recognition model for in-home tasks by training a large amount of data collected from the simulator, randomizing the object color, texture, light conditions, and pose.
Therefore, we also leverage the environment for learning flexible manipulation policies through deep reinforcement learning, and quick testing of the system.

\subsection{Imitation Learning for Manipulation}
We are developing a system with motion planning modules incorporating \textit{imitation learning} to improve generalizability and robustness to environmental changes. We collect expert data for \textit{imitation learning} through teleoperation, in which a person directly teaches the robot how to perform actions using a controller.
We have already successfully adapted a policy to another environment, which differs from where the policy was acquired.
In addition, unlike other motion planning methods, humans teach motions to a robot directly so that a robot may make fast and smooth grasping movements like a human.

\vspace{-0.5\baselineskip}
\section{Contribution}

\noindent\textbf{Re-usability of the System for Other Research Groups}\par
The source codes we developed, including simulators and educational content, are available on GitHub~(\url{https://github.com/matsuolab}).
We also aim to build a platform that integrates and manages data collected from each site belonging to the RoboCup@Home community through real robots and simulators, utilizing our experiences as the AI/ML lab.
The unique characteristics of the RoboCup@Home community are desirable for realizing this platform, for the same robot (HSR) and the similar environment (RoboCup@Home) are located at different locations all over the world, which is one of the reasons why we are eager to take part in DSPL league.
The accumulated data are accessible and usable from each site. Fig.~\ref{fig:Data} shows its conceptual diagram. We will open source the platform and we believe this scheme will greatly accelerate research over the whole community.

\noindent\textbf{Contribution to Expand the Community}\par
We have been sharing our knowledge and findings widely inside and outside the University of Tokyo, regardless of age or affiliation. The aim is to encourage people who do not major in robotics, such as those who specialize in machine learning and data science (we are from the AI/ML lab), to enter robotics by providing them with opportunities, which we believe will lead to the expansion of the community in the long run. Indeed, our activities motivated some members to participate in the RoboCup@Home competition.

\noindent\textbf{Release of the Knowledge Widely}\par
For educational purposes, we open our activities, research achievements, and findings to the public (including junior high and high school students, other universities, and adults). Some of the contents ~\footnote{\url{https://deeplearning.jp/}}  are shown in Table~\ref{table:education}.

\begin{table}[tb]
  \caption{Part of our educational content open to the public}
  \vspace{-0.6\baselineskip}
  \label{table:education}
  \centering
  \begin{tabular}{c|c|c}
    \hline
       Educational Contents & Started & Remarks\\
    \hline \hline
    Data Science Basic Courses& 2014 & more than 5000 participants\\
    \hline
    Deep Learning Basic Courses & 2015 & more than 6000 participants\\
    \hline
    World Models Seminar& 2021 & more than 300 participants\\
    \hline
    Robot System Tutorial~\footnotemark & 2022 & Published on the website\\
    \hline
  \end{tabular}

\end{table}
\footnotetext{\url{https://matsuolab.github.io/roomba_hack_course/course/}}

\noindent\textbf{Research Projects at The University of Tokyo}\par
We have been offering a part of our activities as project-based programs in the Faculty of Engineering at The University of Tokyo since 2021, as opportunities to operate real robots offline besides the online tutorial described in Table~\ref{table:education}.

\vspace{-0.5\baselineskip}
\bibliographystyle{abbrv}
\bibliography{reference}

\newpage
\input{RobotDescriptionDSPL}

\nocite{*}

\end{document}

%% file: RobotDescriptionDSPL.tex
\section*{Software and External Devices}
\label{sec:annex-DSPL}

\setlength\intextsep{0pt}
\begin{wrapfigure}[10]{r}{0.5\textwidth}
	\centering
    \includegraphics[width=5cm]{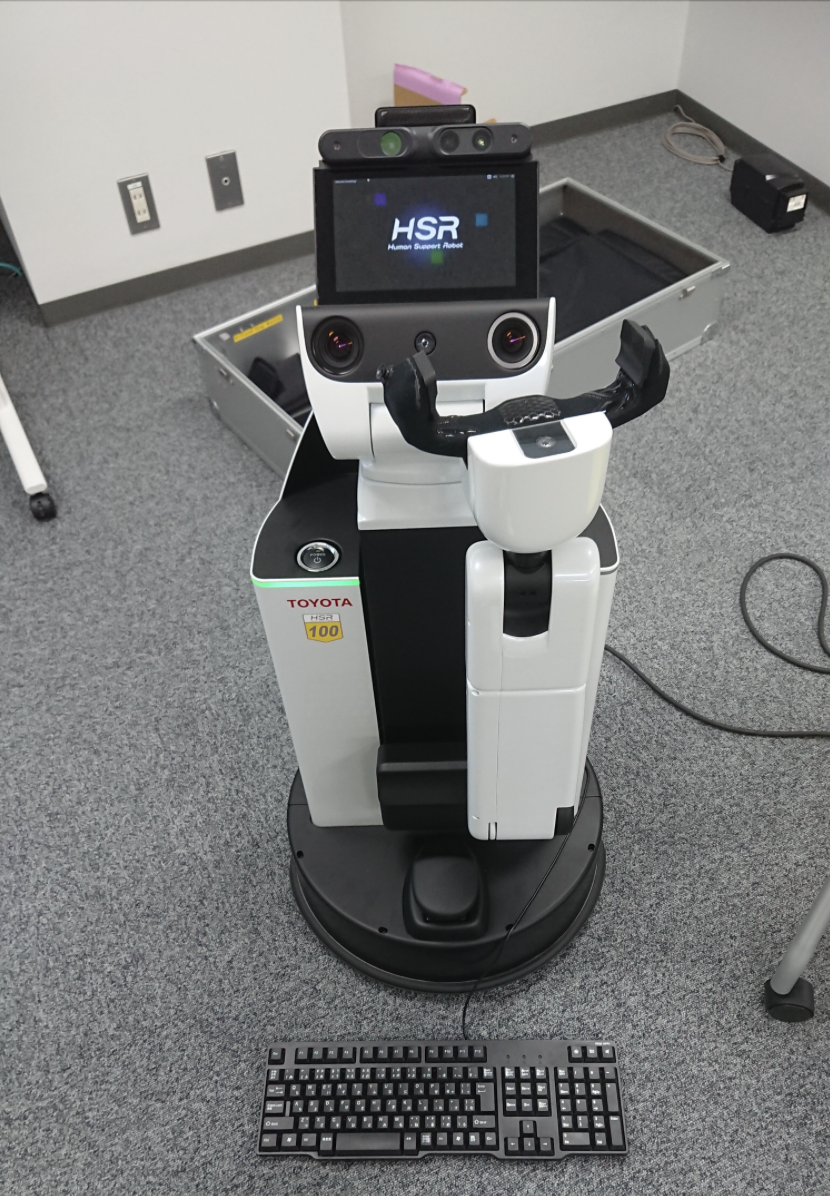}
	\caption{Our HSR}
	\label{fig:eva}
\end{wrapfigure}

We use the \textit{Toyota HSR}. No modifications have been applied.

\section*{Robot's Software Description}

\textit{For our robot we are using the following software:}

\begin{itemize}
	\item Object Detection:  Mask R-CNN, UOIS%
    \item Object Classification:  CLIP (ViT-B/32)
    \item Speech Recognition:  Whisper
    \item NLP and NLU:  GPT-3
    \item Grasping:  Multi-Object Multi-Grasp, GraspNet, GGCNN
    \item Simulators:  Gazebo, PyBullet, Isaac Sim

\end{itemize}

\section*{External Devices}

\textit{Our robot relies on the following external hardware:}

\begin{itemize}
	\item  msi GS66 STEALTH with 10th Gen. Intel® Core™ i9 processor and NVIDIA® GeForce RTX™ 3080 Laptop GPU 16GB GDDR6 mounted on back of HSR
	\item  Desktop PC with AMD Ryzen Threadripper 3970X 32-Core Processor CPU and 2 NVIDIA RTX 3090 GPU connected via wireless network
\end{itemize}